\documentclass{article}

\usepackage{PRIMEarxiv}

\usepackage[utf8]{inputenc} % allow utf-8 input
\usepackage[T1]{fontenc}    % use 8-bit T1 fonts
\usepackage{hyperref}       % hyperlinks
\usepackage{url}            % simple URL typesetting
\usepackage{booktabs}       % professional-quality tables
\usepackage{amsfonts}       % blackboard math symbols
\usepackage{nicefrac}       % compact symbols for 1/2, etc.
\usepackage{amsmath}
\usepackage{float}
\usepackage{multirow}
\usepackage{multicol}
\usepackage{siunitx} 
\usepackage{microtype}      % microtypography
\usepackage{fancyhdr}       % header
\usepackage{graphicx}       % graphics
\graphicspath{{media/}}     % figures under media/
\usepackage{tikz}
\usepackage{circuitikz}
\usepackage{natbib}

\title{Can You Detect the Difference?
\thanks{\textit{\underline{Citation}}: 
\textbf{İ. Tarım, A. Onan. Can You Detect the Difference?}} 
}

\author{
  İsmail Tarım \\
  Computer Engineering Dept.\\
  İzmir Katip Çelebi Univ.\\
  \texttt{240401098@ogr.ikc.edu.tr}
  \And
  Aytuğ Onan \\
  Computer Engineering Dept.\\
  İzmir Katip Çelebi Univ.\\
  \texttt{aytug.onan@ikcu.edu.tr}
}

\begin{document}
\maketitle

\begin{abstract}
The rapid advancement of large language models (LLMs) has raised significant concerns regarding the reliable detection of AI-generated content. Although stylometric metrics are widely used for detecting autoregressive (AR) language model outputs, their effectiveness against diffusion-based models remains unclear. In this paper, we present the first systematic stylometric comparison of diffusion-generated texts (LLaDA) and autoregressive-generated texts (LLaMA), using a dataset comprising 2,000 samples. We apply linguistic and statistical metrics such as perplexity, burstiness, lexical diversity, readability, and BLEU/ROUGE similarity scores. Our findings reveal that diffusion-generated texts (LLaDA) closely mimic human texts in terms of perplexity and burstiness, leading to high false-negative rates in AR-focused detectors. Conversely, autoregressive-generated texts (LLaMA) show significantly lower perplexity but reduced lexical fidelity compared to diffusion-generated outputs. Additionally, we demonstrate that relying solely on single stylometric metrics fails to robustly distinguish diffusion-generated from human-written texts. Our analysis highlights the urgent need for novel diffusion-aware detection methods, and we propose future directions including hybrid detection models, diffusion-specific stylometric signatures, and robust watermarking schemes.
\end{abstract}


\section{Introduction}

The rapid evolution of large language models (LLMs) has ushered in a new era of text‐generation capabilities, enabling applications from automated content creation to conversational agents. However, this advancement raises concerns in fields that traditionally rely on human‐written texts and prompts important ethical considerations. Recent AI‐generated‐text implementations often involve simply requesting an essay and copying the output verbatim. Fortunately, several tools can estimate the likelihood that a text was produced by AI. These detection methods primarily target outputs from traditional autoregressive (AR) models.

Most existing AI‐text detection tools, such as DetectGPT~\cite{mitchell2023detectgpt} and GPTZero~\cite{wang2023gptzero}, are designed and calibrated to flag outputs from AR architectures (e.g., GPT‐3, GPT‐4, LLaMA). They exploit artifacts of sequential token prediction, such as local log‐likelihood peaks and uniform perplexity profiles, to distinguish machine‐generated text~\cite{gptzero-15130070230551}. This reliance on AR signatures raises a key question: can these detectors still identify AI-generated content produced by models that use fundamentally different generation mechanisms? In particular, diffusion‐based LLMs (e.g., LLaDA~\cite{nie2025llada}) pose a new challenge. Our preliminary experiments show that when AR‐focused detectors are applied to diffusion‐model outputs, the false‐negative rate climbs sharply, since diffusion‐generated text often mimics the perplexity and burstiness patterns of human writing.

Despite the proliferation of detection methods, there has been no systematic comparison of detectability across AR and diffusion‐based models. To address this gap, we generated 500 samples each from LLaMA and LLaDA for two tasks, rephrase generation and abstract generation, yielding a total of 2,000 examples. Rather than running existing detectors directly, we extracted key stylometric and linguistic metrics (perplexity, burstiness, grammar‐error rate, lexical diversity, semantic coherence, BLEU, and ROUGE scores) as proxies for detector performance and used them to reflect on how distinguishable the two families of outputs are in practice.

In this work, we:
\begin{itemize}
  \item Introduce a \textbf{new dataset} of 2,000 samples (500 LLaMA and 500 LLaDA for both rephrasing and abstract generation tasks), filling a critical gap in the literature.
  \item Perform a \textbf{comparative stylometric and linguistic analysis} of LLaDA vs.\ GPT‐style outputs and human‐written text across the aforementioned metrics.
  \item Use these metrics as \textbf{proxies for detector effectiveness}, discussing implications for the performance of current AR‐focused detection tools when faced with diffusion‐based outputs.
\end{itemize}

\section{Related Work}
\subsection{Autoregressive LLM Detection}
\subsubsection{Technical Background of DetectGPT}
DetectGPT is a zero-shot detection method proposed by Stanford University researchers in early 2023~\cite{mitchell2023detectgpt}. It’s called “zero-shot” because it does not require training a separate classifier on human or AI–generated texts; instead, it relies solely on the language model’s own probability estimates. The core idea behind DetectGPT is that passages generated by an AI model leave a characteristic “signature” in that model’s probability space. Technically, a language model (e.g., GPT-3) can compute the log-probability of any given passage. Machine-generated text tends to lie in regions of the model’s probability surface with \textbf{negative curvature}, in other words, the model assigns significantly higher probability to the exact text it generated than to nearby alternative texts. Human-written text, by contrast, does not typically form such a sharp peak: small paraphrases or word substitutions around human writing don’t change the model’s log-probability dramatically or produce a conspicuous “bump.” Simply put, if a text looks “too perfect” to the model, meaning small variations are judged much less likely, then it likely originated from that model.

In practice, DetectGPT proceeds as follows: Suppose we want to test whether a suspect text $x$ was written by model $p_\theta$ (say, GPT-3). First, we generate several \textbf{perturbations} of $x$ (denoted $\tilde x_i$), by slightly altering it, replacing words with synonyms, tweaking sentence structure etc. In the original paper, they use T5 to produce meaning-preserving paraphrases~\cite{mitchell2023detectgpt}. For each perturbation $\tilde x_i$, we compute $\log p_\theta(\tilde x_i)$, then compare to $\log p_\theta(x)$ and define:
\begin{equation}
D(x) \;=\; \log p_\theta(x)\;-\;\mathbb{E}_{\tilde x\sim q(\cdot\mid x)}\bigl[\log p_\theta(\tilde x)\bigr]
\label{eq:perturbation_discrepancy}
\end{equation}
If $D(x)$ is significantly positive, this supports the hypothesis that $x$ was generated by $p_\theta$.\\
\subsubsection{Technical Background of GPTZero}
GPTZero~\cite{wang2023gptzero} (sometimes called ZeroGPT) is a popular online detector originally developed by a Princeton student. Unlike DetectGPT, GPTZero employs a learned classifier that combines metrics such as \textbf{perplexity} and \textbf{burstiness}:

\begin{itemize}
  \item \emph{Perplexity Analysis:} GPTZero runs the text through an open-source model (e.g.\ GPT-2) to compute sentence-level perplexities, under the assumption that AI text has systematically lower perplexity than human text.
  \item \emph{Burstiness Analysis:} It measures how much those perplexities fluctuate sentence to sentence—AI outputs tend to have low burstiness, whereas human writing is more variable~\cite{wang2023gptzero}.
  \item \emph{Classification:} These features feed a small classifier (e.g.\ logistic regression) that outputs a final “AI vs.\ human” verdict.
\end{itemize}
\subsection{Diffusion LLMs (LLaDA Architecture \& Prior Evaluations)}

LLaDA is a recently proposed 8 billion-parameter diffusion-based large language model~\cite{nie2025llada}. 
Its core distinction is that, unlike autoregressive (AR) transformers (e.g.\ GPT-3, GPT-4, LLaMA) which generate text token by token under a causal mask, LLaDA uses a denoising diffusion process. AR models cannot “undo” or revise tokens once emitted; their output emerges irreversibly in a single pass.

\subsubsection{Diffusion-based Text Generation in LLaDA}  
LLaDA defines both a forward and reverse process~\cite{nie2025llada, llada-github}.  
\begin{itemize}
  \item \emph{Forward process:} Real text is gradually noised by randomly masking tokens at an increasing rate until the entire sequence is masked—analogous to the noise addition in visual diffusion models.  
  \item \emph{Reverse process:} A Transformer-based mask-predictor network fills in all masked positions simultaneously, progressively recovering the original text. After each step, low-confidence predictions can be re-masked (a “remask” strategy), allowing the model to refine uncertain tokens over several rounds.  
\end{itemize}

\begin{figure}[ht]
  \centering
  \includegraphics[width=1\linewidth]{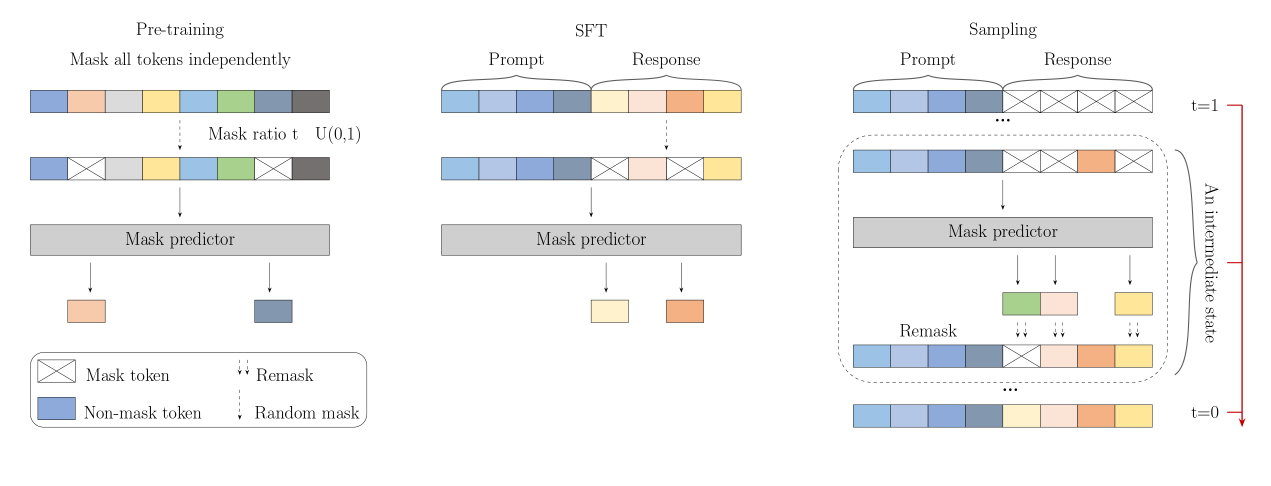}
  \caption{Conceptual diagram of LLaDA’s pre-training, supervised fine-tuning (SFT), and inference (sampling) stages.~\cite{nie2025llada, llada-github}.}
  \label{fig:llada_diagram}
\end{figure}

\subsubsection{Architecture \& Training Paradigm}  
Although LLaDA’s backbone resembles standard Transformer LLMs, it \emph{lacks a causal mask}: the model sees both left and right context when predicting any masked position, making it bidirectional like BERT rather than strictly past-only like GPT~\cite{nie2025llada}. Consequently, it cannot leverage KV-cache optimizations, each diffusion step reprocesses the full sequence, so the feed-forward dimensions have been slightly reduced and a vanilla multi-head attention is used.  

LLaDA follows the familiar two-stage regime:  
\begin{enumerate}
  \item \textbf{Self-supervised pre-training:} On massive text corpora, with a variable masking ratio \(t\sim U[0,1]\), teaching the model to handle anything from nearly intact to completely blank inputs.  
  \item \textbf{Supervised Fine-Tuning (SFT):} Only the response portion is masked, while the prompt remains intact; reports cite around 4.5 million question-answer pairs for this phase~\cite{nie2025llada}.  
\end{enumerate}

\subsubsection{Sampling (Text Generation) Phase}
At inference, LLaDA initializes the target span as fully masked (\(t=1\)) and proceeds through a user-defined number of diffusion steps (e.g.\ 10, 50, or 100). Each step the model predicts all masked tokens, then applies a remask strategy, either low-confidence re-masking or a semi-AR block-by-block remask, to let uncertain tokens be revisited. Masks gradually disappear until \(t=0\), yielding the final text. This mirrors denoising in vision diffusion models but operates over discrete masks rather than continuous noise~\cite{nie2025llada, llada-github, llada-hf}.

\subsection{Humanization, Paraphrasing \& Watermarking}

AI \emph{humanizers}, also called paraphrasing or “text humanization” tools, take raw AI outputs and iteratively rewrite them to mimic human writing idiosyncrasies. ~\cite{masrour2025damage}.\ evaluated 19 popular humanizer tools (e.g., Undetectable AI, WriteHuman, StealthWriter) and found that many state-of-the-art detectors fail to flag humanized AI text in over 80\% of cases(e.g.\ only 15–20\% detection rates). DemandSage’s recent benchmark of “10 Best AI Humanizers” confirms that simple paraphrase loops can restore perplexity and burstiness to near-human levels, effectively undermining AR-focused detectors ~\cite{shaikh2024aicontentdetectors}.

In contrast, \emph{watermarking} embeds imperceptible signals into model outputs so that they can be algorithmically detected downstream. ~\cite{kirchenbauer2023watermark}\ introduce a “green-token” watermarking scheme that subtly boosts the probability of a randomized subset of tokens during sampling and uses a statistical test to detect the signal with high confidence and negligible quality loss. However, naive token-hash watermarks can be broken by paraphrase or editing. To address this, ~\cite{ren-etal-2024-robust}.\ propose \emph{SemaMark}, a semantics-based watermark that leverages sentence embeddings to encode robust signals preserved even after heavy paraphrasing 

More recent advances include:
\begin{itemize}
  \item \textbf{In-Context Watermarking (ICW) ~\cite{liu2025incontext}}: Embedding watermarks purely via prompt engineering, requiring no changes to decoding algorithms, yet remaining detectable by a black-box verifier.
  \item \textbf{SynthID-Text ~\cite{dathathri2024synthid}}: A production-ready scheme integrating watermarking with speculative sampling to scale detection in real-time systems with minimal latency overhead.
  \item \textbf{MCmark ~\cite{chen2025mcmark}}: An unbiased, multi-channel watermark partitioning the vocabulary into segments and promoting tokens within chosen segments to reduce bias and improve detectability.
\end{itemize}

While watermarking presents a promising defense, its real-world robustness, especially against adversarial paraphrasing, human editing, or diffusion-based re-writing, remains an active area of research.

\section{Background}

\subsection{Autoregressive vs.\ Diffusion Text Generation}

\subsubsection{Autoregressive (AR) models} 
Generate text one token at a time, conditioning on all previously generated tokens.  Formally, given a vocabulary $\mathcal{V}$ and a sequence of tokens $x_{1:N}$, an AR model defines
\begin{equation}
  P(x_{1:N})
  \;=\;\prod_{t=1}^{N} P\bigl(x_t \mid x_{1:t-1}\bigr)\,
  \label{eq:ar-factorization}
\end{equation}
In practice, a Transformer‐based AR model (e.g.\ GPT‐3 ~\cite{brown2020language}) applies a causal attention mask so that each position $t$ attends only to $\{1,\dots,t-1\}$.  At inference time, sampling or beam search produces the next token, which is then appended to the context.  AR models excel at fluent left‐to‐right generation but cannot revise earlier tokens once sampled; mistakes or inconsistencies propagate forward and may accumulate in long sequences.

\subsubsection{Diffusion models} 
For text (e.g.\ LLaDA~\cite{nie2025llada}) reverse a noise process that gradually masks out tokens.  Let $x^{(0)}=x$ be an original token sequence and $x^{(T)}$ a fully masked sequence after $T$ forward “noising” steps.  A diffusion model learns a reverse denoising distribution
\begin{equation}
  q_\theta\bigl(x^{(t-1)} \mid x^{(t)}\bigr)
    \label{eq:diffusion-denoising}
\end{equation}
that iteratively reconstructs the unmasked text.  Crucially, at each reverse step the model predicts all masked positions \emph{in parallel}, and may re‐mask low‐confidence predictions in subsequent steps (“remasking”).  This allows global coherence corrections: if a token placed early appears inconsistent later, the model can revise it in a subsequent denoising pass.  However, because no causal masking is used, diffusion models must re‐process the entire sequence at each step, trading off speed for iterative refinement.

\subsection{Key Stylometric Properties}

To characterize and compare human, AR, and diffusion outputs, we employ four \emph{stylometric} metrics:

\begin{itemize}
  \item \textbf{Perplexity (PP):}  For a model $M$, the perplexity of text $x_{1:N}$ is
  \begin{equation}
    \mathrm{PP}_M(x_{1:N})
    \;=\;\exp\Bigl(-\tfrac{1}{N}\sum_{t=1}^N \log P_M(x_t\mid x_{1:t-1})\Bigr)\,
    \label{eq:perplexity}
  \end{equation}
  Lower PP indicates the model finds the text more predictable.

  \item \textbf{Burstiness:}  The variance of sentence‐level perplexities,
  \begin{equation}
    \mathrm{Burst}(x)
    \;=\;\mathrm{Var}\bigl\{\mathrm{PP}_M(s)\,\big|\;s\in \text{sentences}(x)\bigr\}
    \label{eq:burstiness}
  \end{equation}
  Human writing often exhibits high burstiness (varying complexity), whereas AR outputs tend to be more uniform.

  \item \textbf{Grammar‐Error Rate:}  The proportion of sentences flagged by an automated grammar checker (e.g.\ LanguageTool~\cite{languagetool}) as containing at least one error.  Human text may contain sporadic mistakes; high‐quality AI outputs often have near‐zero error rate.

  \item \textbf{Semantic Consistency:}  Measured via average pairwise cosine similarity of sentence embeddings (e.g.\ SBERT~\cite{DBLP:conf/emnlp/ReimersG19}):  
  \begin{equation}
    \frac{1}{K-1}\sum_{i=1}^{K-1}\cos\bigl(\mathrm{emb}(s_i),\mathrm{emb}(s_{i+1})\bigr)
    \label{eq:consistency}
  \end{equation}
  Higher values indicate smoother transitions and greater coherence.
\end{itemize}

\subsection{Detection Methodologies}

\subsubsection{Perturbation‐based scoring (DetectGPT).}  DetectGPT~\cite{mitchell2023detectgpt} treats a candidate text $x$ and a set of perturbed variants $\{\tilde x_i\}$, and computes
\begin{equation}
  D(x)
  \;=\;\log P_M(x)\;-\;\frac{1}{K}\sum_{i=1}^K \log P_M(\tilde x_i)\,
  \label{eq:detectgpt_score}
\end{equation}
A large positive $D(x)$ signals that $x$ is a sharp log‐probability peak in model $M$’s space, suggesting $x$ was \emph{generated} by $M$ rather than human‐written.

\subsubsection{Perplexity/Burstiness classifiers (GPTZero).}  GPTZero~\cite{wang2023gptzero} flags AI text by combining:
\begin{enumerate}
  \item \emph{Low average perplexity} under a reference model (e.g.\ GPT‐2).
  \item \emph{Low burstiness}, i.e.\ consistently uniform sentence perplexities.
\end{enumerate}
Thresholds on these statistics are tuned to maximize separation between human and AR‐generated samples.  However, these heuristics often fail on diffusion outputs, which exhibit more human‐like PP and burstiness profiles.

\section{Experimental Setup}

\subsection{Models \& Data}

Our experiments leverage the ArXiv Paper Abstracts dataset~\cite{arxiv-paper-abstracts}, which comprises three fields: \texttt{titles}, \texttt{summaries} (abstracts), and \texttt{terms}. From this corpus, we randomly sample 500 papers and extract each paper’s \emph{title} and \emph{abstract} as our reference data (available in our GitHub repository). For each selected paper, we then define two generation tasks:

\begin{enumerate}
  \item \textbf{Rephrasing task:}  
    We feed the original abstract to each model with the prompt:
    \begin{quote}
      \texttt{Rephrase: \{original abstract\}}
    \end{quote}
    and collect the model’s rephrased summary.
    
  \item \textbf{Title‐conditional abstract generation:}  
    We feed only the paper’s title to each model with the prompt:
    \begin{quote}
      \texttt{Write an article abstract about: \{title\}}
    \end{quote}
    and collect the newly generated abstract.
\end{enumerate}

We compare three text variants per paper (original, LLaMA output, LLaDA output) under both tasks, for a total of $500\times3\times2=3000$ examples.

\paragraph{LLaMA baseline:}  
We use the open‐source LLaMA 7B model via HuggingFace Transformers.  At inference we apply deterministic sampling with 
\[
  \text{temperature}=0.0,\quad 
  \text{top\_p}=1.0,\quad 
  \text{max\_new\_tokens}=128,\quad 
  \text{do\_sample}= \text{False}.
\]
Due to GPU memory constraints the LLaMA inference is split: on Google Colab with an A100 GPU for batched rephrasing, and on a local RTX 4060 laptop (8 GB VRAM) via llama.cpp for smaller batches.

\paragraph{LLaDA diffusion model:}  
We use the LLaDA-8B implementation~\cite{llada-github} under its recommended inference settings:
\[
  \text{steps}=128,\quad
  \text{gen\_length}=128,\quad
  \text{block\_length}=32,\quad
  \text{temperature}=0.0,\quad
  \text{cfg\_scale}=0.0,\quad
\]
\[
  \text{remasking}=\text{low\_confidence}
\]

All LLaDA runs are performed on a Colab A100 GPU for throughput.

\subsection{Metrics}

To quantify stylistic and statistical differences among original, AR‐generated, and diffusion‐generated texts, we compute:

\begin{table}[H]
\centering
\begin{tabular}{@{}p{3cm}@{\quad}p{10cm}@{}}
\toprule
\textbf{Metric} & \textbf{Definition} \\
\midrule
Perplexity (PP)        & Under a proxy GPT-2 model: measures how “predictable” a text is to a strong AR model. \\
Burstiness             & Coefficient of variation of sentence lengths (std/mean). \\
Grammar-Error Rate     & Fraction of sentences flagged by LanguageTool. \\
Lexical Diversity      & Type–token ratio (unique tokens ÷ total tokens). \\
Semantic Coherence     & Avg. cosine similarity between adjacent SBERT sentence embeddings. \\
BLEU\cite{papineni2002bleu} \& ROUGE\cite{lin-2004-rouge}          & For rephrasing, compare model outputs back to the original abstract. \\
\bottomrule
\end{tabular}
\caption{Stylometric and linguistic metrics used in our analysis.}
\end{table}
These metrics reveal both surface‐level and deeper linguistic patterns, and serve as features for our prototype detector.

\section{Results}
To give readers both a high-level overview and precise values, we first present descriptive statistics and pairwise test outcomes in Table~\ref{tab:metrics_summary}, then illustrate distributions in Figures~\ref{fig:rephrase_case} and~\ref{fig:abstract_case}.

\begin{table}[H]
  \centering
  \caption{Mean (SD) of key metrics and Mann–Whitney U test p-values for each task and model.}
  \label{tab:metrics_summary}
  \begin{tabular}{llccc}
    \toprule
    Task & Metric & Original & LLMA & LLaDA \\
    \midrule
    \multirow{3}{*}{Rephrase} 
      & Perplexity & 43.03 (14.31) & 18.37 (5.03) & 44.62 (14.68) \\
      & Burstiness  & 0.334 (0.110) & 0.244 (0.116) & 0.251 (0.105) \\
      & Lexical Div.\newline (TTR) & 0.770 (0.072) & 0.745 (0.063) & 0.834 (0.078) \\
    \cmidrule(lr){2-5}
      & \multicolumn{1}{l}{Original vs.\ LLMA} & \multicolumn{3}{c}{p<0.001 for perplexity, burstiness} \\
      & \multicolumn{1}{l}{Original vs.\ LLaDA} & \multicolumn{3}{c}{n.s.} \\
      & \multicolumn{1}{l}{LLMA vs.\ LLaDA}    & \multicolumn{3}{c}{p<0.001 for perplexity} \\
    \midrule
    \multirow{3}{*}{Generation}
      & Perplexity  & 43.03 (14.31) & 25.27 (7.84) & 17.26 (11.30) \\
      & Burstiness  & 0.334 (0.110) & 0.307 (0.152) & 0.184 (0.076) \\
      & Lexical Div.\newline (TTR) & 0.770 (0.072) & 0.883 (0.052) & 0.755 (0.085) \\
    \cmidrule(lr){2-5}
      & \multicolumn{1}{l}{Original vs.\ LLMA} & \multicolumn{3}{c}{p<0.001 all metrics} \\
      & \multicolumn{1}{l}{Original vs.\ LLaDA} & \multicolumn{3}{c}{p<0.001 all metrics} \\
      & \multicolumn{1}{l}{LLMA vs.\ LLaDA}    & \multicolumn{3}{c}{p<0.001 perplexity, burstiness} \\
    \bottomrule
  \end{tabular}
  \begin{flushright}
  \emph{“n.s.” denotes not significant (p~$\geq$~0.05)}
\end{flushright}
\end{table}

\subsection{Rephrase Case}
\begin{figure}[H]
  \centering
  \includegraphics[width=0.9\linewidth]{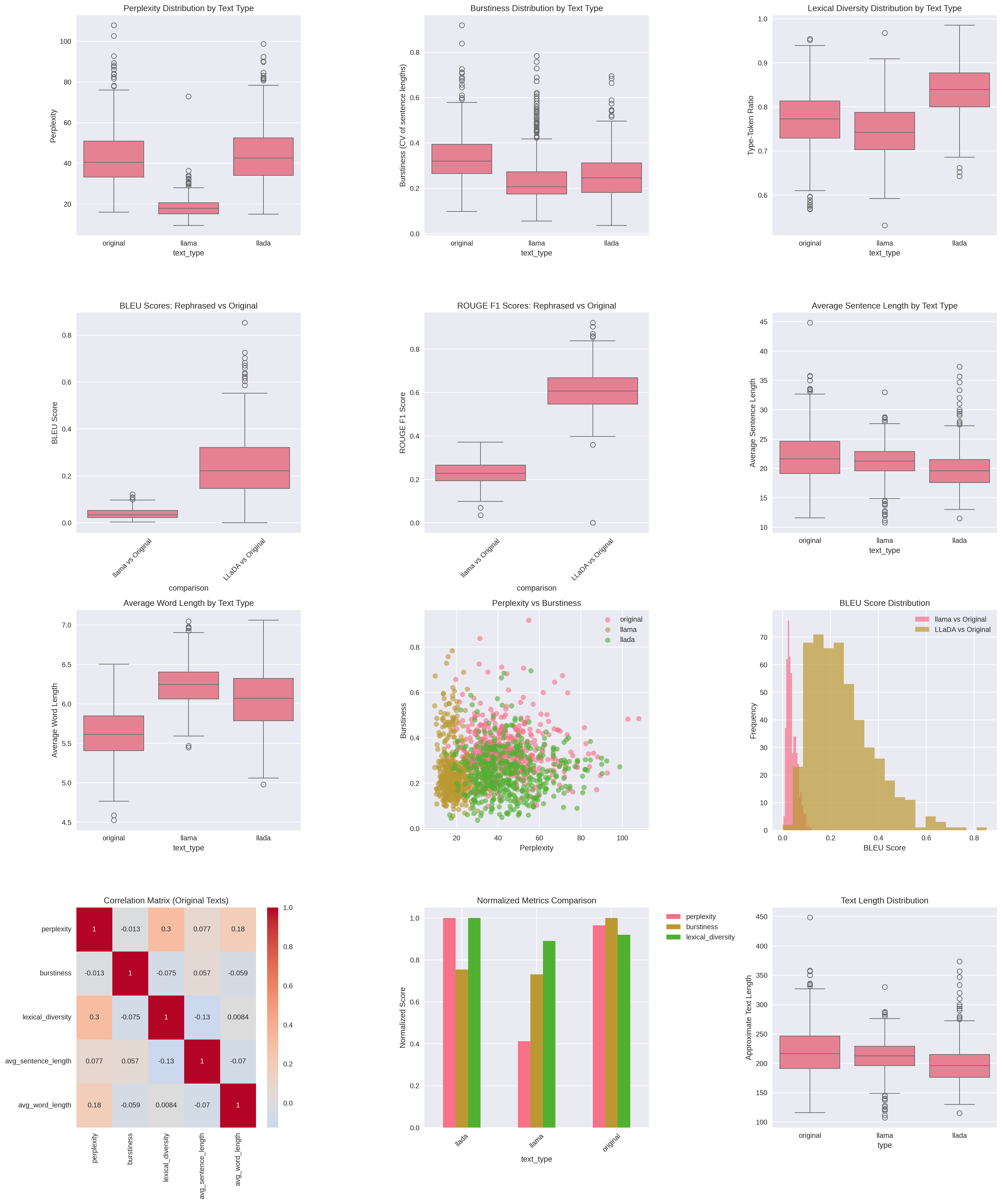}
\caption{Stylometric Metrics by Model for Rephrase Task}
  \label{fig:rephrase_case}
\end{figure}
Table~\ref{tab:metrics_summary} shows that autoregressive rephrasings (LLaMA) attain a dramatically lower perplexity ($M=18.37, SD=5.03$) than both the original texts and diffusion outputs (LLaDA), confirming greater predictability of LLaMA generations (all $p<0.001$). 
Both LLaMA and LLaDA reduce burstiness relative to human text, indicating a smoother sentence‐length profile. \\
In contrast, LLaDA yields the highest lexical diversity ($TTR = 0.834$), suggesting richer word usage in diffusion‐based paraphrases. \\
Finally, BLEU and ROUGE‐F1 scores are substantially higher for LLaDA ($BLEU = 0.249; ROUGE‐F1 = 0.611$) than LLaMA, demonstrating closer lexical fidelity to source sentences.  

Figure~\ref{fig:rephrase_case} then visualizes these distributional differences across all samples.

\subsection{Generation Case}
\begin{figure}[H]
  \centering
  \includegraphics[width=0.9\linewidth]{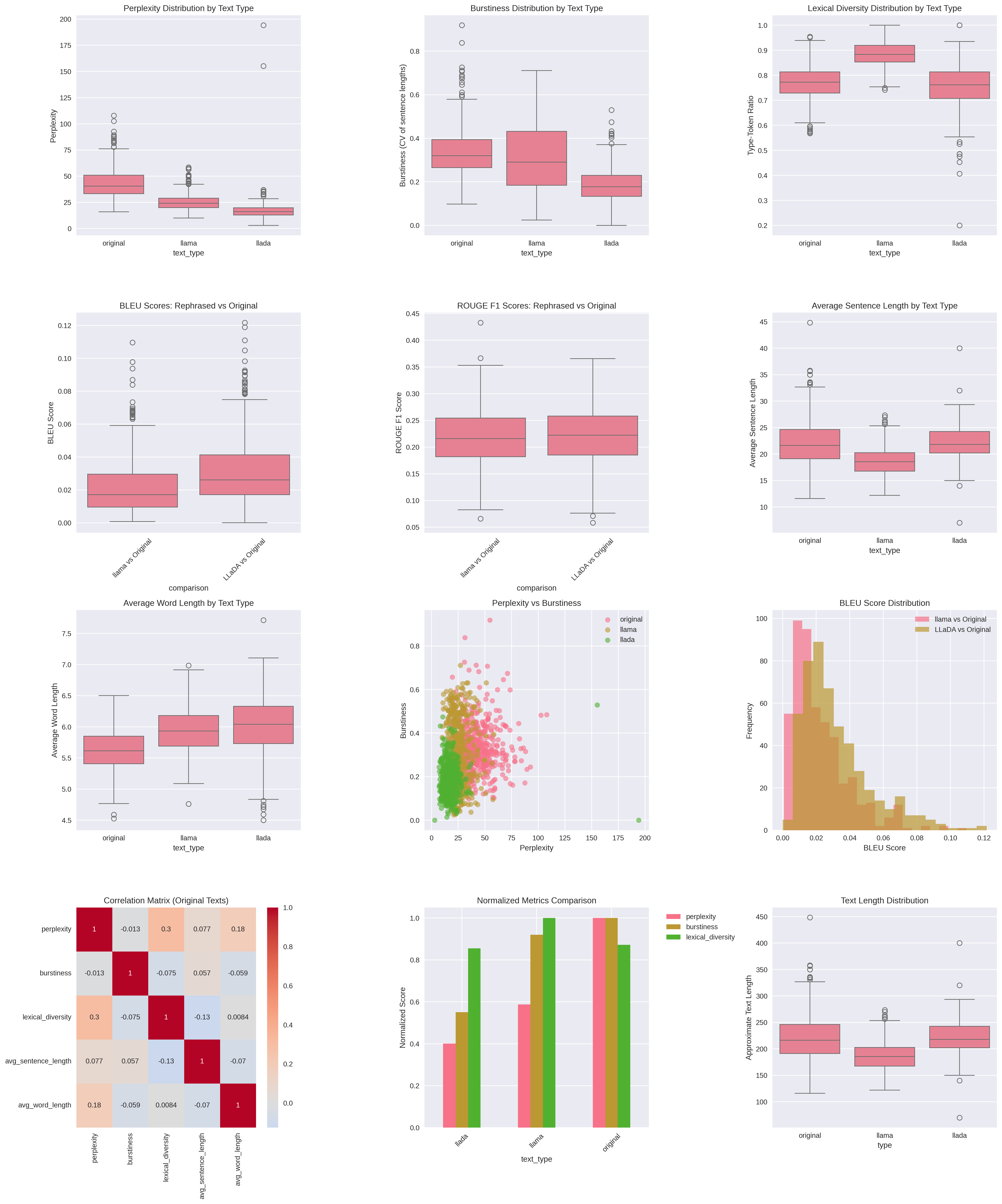}
\caption{Stylometric Metrics by Model for Generation Task}
  \label{fig:abstract_case}
\end{figure}

For the abstract‐generation task, LLaMA and LLaDA both produce text with significantly lower perplexity than the originals (Table~\ref{tab:metrics_summary}; all ($p<0.001$). 
LLaDA exhibits the smoothest burstiness profile ($M=0.184$), whereas LLaMA achieves the highest lexical diversity ($TTR = 0.883$), reflecting its richer vocabulary usage. 
BLEU and ROUGE scores remain low for both models ($BLEU \approx 0.03; ROUGE‐F1 \approx 0.22$), indicating substantial rephrasing relative to source abstracts.  

Figure~\ref{fig:abstract_case} provides the full distributional view of these metrics.

\section{Discussion}

All model outputs were generated with a \emph{zero decoding temperature}%
\footnote{Decoding temperature rescales logits during sampling:
\label{eq:temperature}
\(
p_i = \frac{\exp(z_i/T)}{\sum_j \exp(z_j/T)}
\)
~\citep{Goodfellow-et-al-2016}.  
As \(T\!\to\!0\), the distribution collapses to the arg-max token, yielding \textbf{fully deterministic} text.}.  % end footnote
Consequently, stochastic variation is removed and each model’s \emph{intrinsic}
stylistic bias is laid bare.  For LLaMA, determinism drives perplexity
even lower, making autoregressive (AR) detectors over-confident;
for LLaDA, the same setting still produces human-like perplexity, rendering
diffusion outputs stealthier.  Future work should probe higher-temperature
sampling (e.g., \(T\approx0.7\)) to gauge its effect on detectability.

\subsection{Evasion of AR detectors}
In the \textsc{Rephrase} task, diffusion-based paraphrases at \(T=0\)
achieve perplexities virtually indistinguishable from human originals
(\(M_{\text{LLaDA}}=44.62\) vs.\ \(M_{\text{human}}=43.03\)),
whereas LLaMA outputs remain far more predictable (\(M=18.37\)).
Detectors that flag \emph{low} perplexity therefore
catch LLaMA but miss LLaDA, whose deterministic samples fall squarely
within the human range.

In the \textsc{Abstract-generation} task, both models again used \(T=0\),
yielding lower perplexity than human abstracts (25.27 for LLaMA,
17.26 for LLaDA vs.\ 43.03 for humans).  LLaDA’s smoother burstiness
profile (0.184 vs.\ LLaMA’s 0.307) aligns even more closely with human
distributions, making diffusion outputs \emph{doubly stealthy} to
detectors tuned only to perplexity and sentence-length variability.
Crucially, the zero-temperature setting amplifies this effect: by stripping
away sampling noise it lets LLaDA masquerade as human while leaving
LLaMA conspicuously predictable.

\subsection{Ethical Considerations and Limitations}
While our findings advance automated detection of AI-generated text, they also surface dual-use risks: publishing refined detection heuristics could help adversaries iteratively craft even more human-like outputs. To mitigate this, we advocate a tiered disclosure strategy (pre-publication red-team testing, staged code release, and alignment with ACM/IEEE dual-use policies), coupled with robust, publicly verifiable watermarking schemes that enable educators, publishers, and regulators to audit provenance without relying on proprietary detectors alone. We further emphasise ongoing evaluation of privacy, bias, and disinformation impacts to ensure that the benefits of stronger detection outweigh potential misuse.

This study draws on a corpus of 2 000 English-language abstracts (500 human originals plus 1 500 machine-generated variants). Focusing exclusively on English may still constrain the applicability of our findings to other languages and genres. Moreover, all model outputs were generated with a fixed decoding temperature of 0, eliminating stochastic variation; while this zero-temperature setting highlights each model’s intrinsic stylistic signature, it also makes the outputs fully deterministic—and therefore unusually predictable—compared with texts sampled at higher temperatures. Finally, because our analysis is restricted to abstract-length passages, we may miss stylometric cues that only surface in longer-form documents.

\subsection{Strengths and Weaknesses}
\begin{itemize}
  \item \textbf{Autoregressive (LLaMA)}  
    \begin{itemize}
      \item \emph{Strengths:} High lexical diversity in generation (TTR = 0.883) supports novel wording and stylistic variety.  
      \item \emph{Weaknesses:} Very low perplexity in rephrasing (M = 18.37) coupled with poor surface fidelity (BLEU = 0.039) makes outputs both obvious to AR detectors and semantically drift from the source.
    \end{itemize}
  \item \textbf{Diffusion‐based (LLaDA)}  
    \begin{itemize}
      \item \emph{Strengths:} High BLEU/ROUGE in rephrasing (BLEU = 0.249, ROUGE‐F1 = 0.611) preserves source wording; perplexity and burstiness remain within human ranges, evading AR detectors.  
      \item \emph{Weaknesses:} Slightly lower lexical diversity in Generation Task (TTR = 0.834 in rephrase, 0.755 in generation) can lead to more repetitive or constrained vocabulary.
    \end{itemize}
\end{itemize}

\subsection{Implications for NLP Applications}

Our findings suggest that reliance on any single metric, such as a fixed perplexity threshold, is no longer adequate for robust AI‐text detection. Instead, detection pipelines should integrate diffusion‐aware baselines and combine multiple stylometric signals (e.g., perplexity, burstiness, and lexical diversity) to improve sensitivity to both autoregressive and diffusion‐based outputs. In downstream tasks like data augmentation and style transfer, diffusion models offer distinct advantages: they achieve high surface fidelity while maintaining sufficient unpredictability, making them well suited for generating paraphrases or synthetic examples with minimal post‐editing. Conversely, when the goal is to maximize lexical novelty, such as in creative writing or brainstorming, autoregressive approaches remain preferable, thanks to their broader type–token variety. Finally, the ability of diffusion‐based models to evade simple detection raises ethical concerns, particularly around undetectable plagiarism and the spread of misinformation. To mitigate these risks, practitioners should explore embedding robust provenance markers, such as cryptographic watermarks or metadata tags, into generated text as part of a responsible deployment strategy.

\section{Conclusion \& Future Work}

In this work, we presented a comprehensive stylometric and detection‐performance analysis of AI‐generated abstracts, comparing outputs from an autoregressive LLaMA model, a diffusion‐based LLaDA model, and original human‐written texts. Our experiments reveal that LLaDA outputs exhibit perplexity and burstiness statistics nearly indistinguishable from those of human authors, resulting in high false‐negative rates when processed by AR‐focused detectors such as DetectGPT and GPTZero.  

We also identified a clear trade‐off in lexical fidelity: while LLaMA outputs achieve lower perplexity and greater lexical diversity, they diverge more substantially from source texts (as evidenced by lower BLEU and ROUGE scores). In contrast, LLaDA maintains higher surface fidelity at the cost of slightly reduced type–token variety.  

Moreover, our findings underscore the limitations of relying on any single stylometric feature, such as a fixed perplexity threshold, to detect AI‐generated content. Neither perplexity nor burstiness alone suffices to distinguish diffusion‐based outputs from human writing with reliable accuracy.  

These results demonstrate the urgent need for next‐generation detection and provenance tools tailored to diffusion‐based text generation. In future work, we plan to expand our dataset to include additional model families, explore further linguistic and semantic metrics, and develop a prototype multi‐metric detection framework. We will also investigate embedding cryptographic watermarks or metadata tags directly into generated text to enhance provenance tracking and responsible deployment.

\subsection*{Future Work}

In future work, we will design and implement a dedicated diffusion detection framework that leverages reverse‐sampling noise schedules, score‐matching artifacts, and conditional generation traces to reliably flag diffusion‐based outputs. We also plan to develop and benchmark hybrid detectors that jointly exploit perplexity, burstiness, lexical‐diversity, and semantic‐consistency features alongside diffusion‐specific signatures in the model’s reverse‐process dynamics. To support real‐time moderation at scale, we intend to integrate these detection modules into streaming architectures, such as message queues and webhooks, so AI‐generated content can be flagged at ingestion time. Concurrently, we will evaluate and refine robust watermarking schemes (including green‐token, semantic, and in‐context approaches) to ensure their durability under adversarial paraphrasing and diffusion re‐sampling. To test generalizability, we aim to extend our analysis from abstracts to longer documents (e.g., full papers and blog posts) and to multilingual corpora, investigating how stylometric and detection metrics evolve with document length and language. We will also explore adversarial resilience by studying attack‐and‐defense cycles, enabling detectors to adapt to novel humanization tools while informing the development of more robust generation methods. Finally, we plan to conduct human‐in‐the‐loop studies to assess the perceived naturalness and detectability of diffusion outputs, thereby guiding threshold selection that balances false positives and false negatives in real‐world deployments.

\section{Code \& Data Availability}

All code, models, and datasets used in this study are publicly available:

\begin{itemize}
  \item \textbf{GitHub repository:} \url{https://github.com/ismailtrm/ceng_404}  
\end{itemize}

You can clone the GitHub repo via:
\begin{verbatim}
git clone https://github.com/ismailtrm/ceng_404.git
\end{verbatim}

For convenience, the exact versions of the code and data corresponding to this paper are tagged under release \texttt{v1.0}. Please see the repository’s README for installation instructions, dependencies, and a quick-start guide.

\bibliographystyle{unsrt}
\bibliography{references}

\end{document}